\documentclass{article}

\usepackage{iclr2027_conference,times}

\newif\ifpreprint
\preprinttrue
\ifpreprint
\iclrfinalcopy
\fi

\usepackage[utf8]{inputenc}
\DeclareUnicodeCharacter{03C4}{\ensuremath{\tau}} %
\usepackage[T1]{fontenc}
\usepackage{hyperref}
\usepackage{url}
\usepackage{booktabs}
\usepackage{array}
\usepackage{listings}
\usepackage{amsfonts}
\usepackage{nicefrac}
\usepackage[patch=none]{microtype}
\usepackage{xcolor}
\definecolor{posgreen}{HTML}{009E73}
\definecolor{negred}{HTML}{D55E00}
\usepackage{graphicx}
\usepackage{subcaption}
\usepackage[textsize=footnotesize,colorinlistoftodos,textwidth=3.2cm,prependcaption]{todonotes}
\usepackage{enumitem}
\usepackage{booktabs}
\usepackage{xcolor}
\usepackage{pifont}
\usepackage{array}

\usepackage{amsmath,amsfonts,bm}

\def\figref#1{\hyperref[#1]{Figure~\ref*{#1}}}
\def\Figref#1{\hyperref[#1]{Figure~\ref*{#1}}}
\def\twofigref#1#2{Figures \hyperref[#1]{\ref*{#1}} and \hyperref[#2]{\ref*{#2}}}
\def\quadfigref#1#2#3#4{Figures \hyperref[#1]{\ref*{#1}}, \hyperref[#2]{\ref*{#2}}, \hyperref[#3]{\ref*{#3}} and \hyperref[#4]{\ref*{#4}}}
\def\secref#1{\hyperref[#1]{Section~\ref*{#1}}}
\def\Secref#1{\hyperref[#1]{Section~\ref*{#1}}}

\def\eqref#1{equation~\ref{#1}}

\def\1{\bm{1}}

\DeclareMathAlphabet{\mathsfit}{\encodingdefault}{\sfdefault}{m}{sl}
\SetMathAlphabet{\mathsfit}{bold}{\encodingdefault}{\sfdefault}{bx}{n}

\newif\ifsubmission
\submissiontrue %

\newcommand{\cmark}{\textcolor{green!70!black}{\ding{51}}}
\newcommand{\xmark}{\textcolor{red}{\ding{55}}}

\ifsubmission
 \newcommand{\xhtodo}[1]{}
 \newcommand{\fptodo}[1]{}
 \newcommand{\pbtodo}[1]{}
 \newcommand{\ahtodo}[1]{}
 \newcommand{\dvtodo}[1]{}
 \newcommand{\generaltodo}[1]{}
\else
 \newcommand{\xhtodo}[1]{\todo[caption={XH},color=blue!25,linecolor=blue!60,bordercolor=blue!60]{#1}}
 \newcommand{\fptodo}[1]{\todo[caption={FP},color=red!20,linecolor=red!60,bordercolor=red!60]{#1}}
 \newcommand{\pbtodo}[1]{\todo[caption={PB},color=green!25,linecolor=green!60,bordercolor=green!60]{#1}}
 \newcommand{\ahtodo}[1]{\todo[caption={AH},color=yellow!35,linecolor=yellow!70,bordercolor=yellow!70]{#1}}
 \newcommand{\dvtodo}[1]{\todo[caption={DV},color=violet!25,linecolor=violet!60,bordercolor=violet!60]{#1}}
 \newcommand{\generaltodo}[1]{\todo[caption={TODO},color=orange!30,linecolor=orange!70,bordercolor=orange!70]{#1}}
\fi

\newcommand{\pnnum}[1]{#1} %

\newcommand{\bucketLow}{\textsc{AgentHorizon-Simple}}
\newcommand{\bucketHigh}{\textsc{AgentHorizon}}
\newcommand{\bucketDev}{\textsc{AgentHorizon-Development}}

\newcommand{\ahReviewedCount}{\pnnum{1{,}373}}
\newcommand{\ahDevelopmentCount}{\pnnum{162}}
\newcommand{\ahEvaluationCount}{\pnnum{1{,}211}}
\newcommand{\ahFinalCount}{\pnnum{528}}
\newcommand{\ahSimpleFinalCount}{\pnnum{683}}

\providecommand{\ahKeyframe}[3]{%
 \begin{minipage}[t]{0.245\linewidth}%
 \centering
 \includegraphics[width=\linewidth]{#1}\\[0.15em]
 {\scriptsize\textbf{#2}~\textit{#3}}%
 \end{minipage}%
}

\title{AgentHorizon: Evaluating Agentic Judges for Long-Horizon Computer-Use Tasks}

\author{%
\textbf{Xing Han L\`{u}\textsuperscript{1,2,3}\quad Dheeraj Vattikonda\textsuperscript{2,3,\ensuremath{\dagger}}\quad Sina Hajimiri\textsuperscript{4,\ensuremath{\dagger}}}\\
\textbf{Fatemeh Pesaran Zadeh\textsuperscript{5}\quad Parishad BehnamGhader\textsuperscript{2,3,\ensuremath{\dagger}}\quad Ghazwa Darwiche\textsuperscript{1}}\\
\textbf{Amirhossein Kazemnejad\textsuperscript{3}\enspace Christopher Pal\textsuperscript{1,3,7,8}\enspace Alexandre Drouin\textsuperscript{1,3,6}\enspace Siva Reddy\textsuperscript{2,3,\ensuremath{\dagger}}}\\[0.4em]
\normalfont\small\textsuperscript{1}ServiceNow Research\quad\textsuperscript{2}McGill University\quad\textsuperscript{3}Mila -- Quebec AI Institute\\
\normalfont\small\textsuperscript{4}\'{E}TS Montr\'{e}al\quad\textsuperscript{5}Seoul National University\quad\textsuperscript{6}Universit\'{e} Laval\\
\normalfont\small\textsuperscript{7}Polytechnique Montr\'{e}al\quad\textsuperscript{8}Canada CIFAR AI Chair\\[0.2em]
\normalfont\small\textsuperscript{\ensuremath{\dagger}}Work done while at ServiceNow Research.
}

\begin{document}
\raggedbottom

\maketitle
\ifpreprint\lhead{Preprint}\fi
\suppressfloats[t]

\begin{abstract}
Computer-use agents are capable of completing complex tasks, increasing the use of automatic judges to determine success, either for training or for evaluation without human involvement. Despite their flexibility, their reliability on long tasks spanning multiple applications remains unclear. A trajectory, composed of long sequences of screenshots and actions, may appear complete, but in reality violates constraints from the instruction or introduces an unwanted side effect. To identify these errors, a judge needs to carefully examine the trajectory with respect to the user's instruction. To this end, we introduce \textsc{AgentHorizon}, a benchmark of $\pnnum{1{,}373}$ computer-use tasks (instruction--trajectory pairs) drawn from $\pnnum{166}$ hours of human-recorded trajectories spanning three operating systems. By recording trajectories for closely related instructions, we can construct negative tasks by simply \textit{swapping} the instructions. This paired design evaluates judges on their ability to distinguish a truly successful trajectory from one that completed a similar (but incompatible) request.
We release the benchmark under three splits: a frontier split, \textsc{AgentHorizon} (AH), a simplified split, \textsc{AgentHorizon-Simple} (AH-S), and a development split, \textsc{AgentHorizon-Development} (AH-D). We further evaluate eleven judges by (1) directly passing the full trajectory (with up to 300 screenshots and actions), and (2) using them as coding agents across five agent harnesses. We find that our best agentic judge, GPT-5.5, achieves $\pnnum{80.9\%}$ balanced accuracy on the AH subset. We find that tool-use improves certain models but results in worse performance for open-weight models, and that judges differ drastically in their ability to accept a valid trajectory and reject failed ones. Our findings highlight the need for judges that are capable of locating and verifying often hidden evidence that a task was properly completed inside long interaction histories. We release our benchmark and results publicly.
\end{abstract}

\ificlrfinal
\begin{center}
\small\url{https://github.com/ServiceNow/agenthorizon}
\end{center}
\fi

\section{Introduction}
\label{sec:intro}
\begin{figure}[t]
 \centering
 \includegraphics[width=\linewidth]{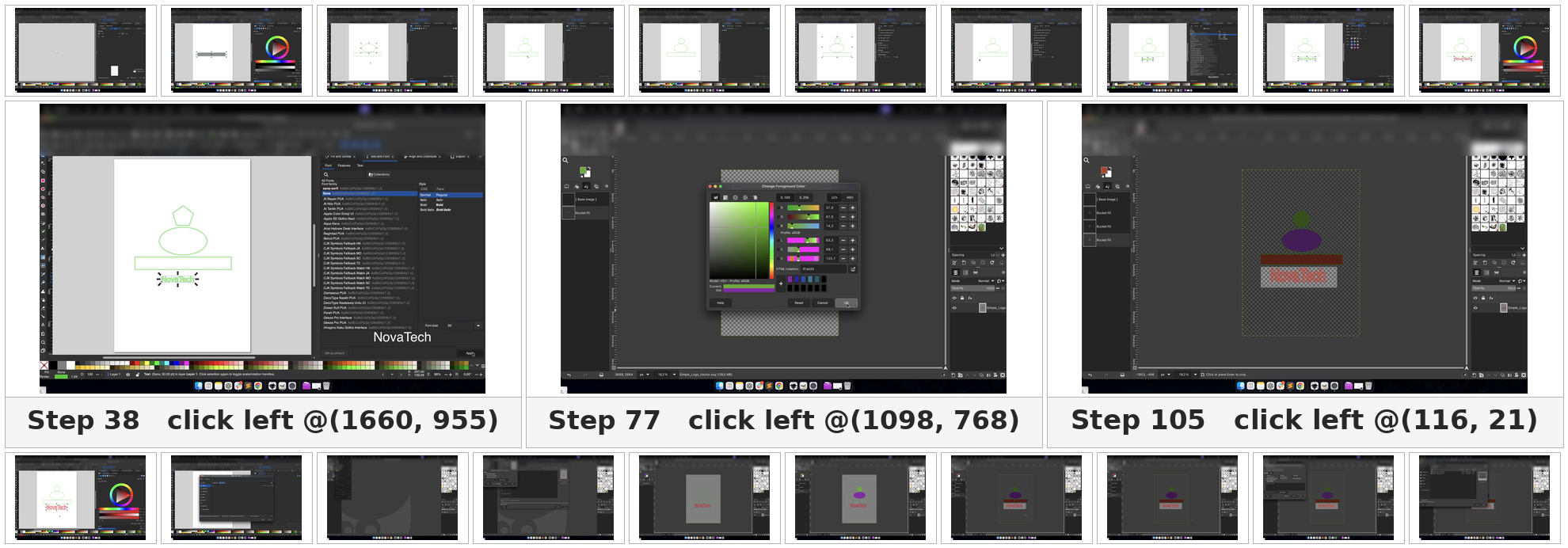}
 \caption{A plausible output can violate the instruction. In this $\pnnum{115}$-step Inkscape--GIMP recording (\texttt{0cacdac6}, macOS), a pentagon replaces the requested triangular accent. The mismatch survives recoloring and export, giving the crossed item a Critical Mistake label. The outer strips sample $\pnnum{10}$ frames from each half of the recording; the enlarged frames show the canvas, color selection, and final composite at steps $\pnnum{38}$, $\pnnum{77}$, and $\pnnum{105}$.}
 \label{fig:trajectory-progression}
\end{figure}

Recent computer-use agents are becoming increasingly capable of carrying out complex tasks through graphical interfaces, from editing a spreadsheet to preparing a design across several desktop applications. However, measuring their progress requires determining if a sequence of actions, or \emph{trajectory}, correctly fulfills user requests. For longer workflows, this may require verifying details scattered across hundreds of screenshots and action logs; we need to identify whether an item was properly selected, whether a sensitive file was shared with the intended recipients, or whether a design was created with the valid specifications.

Existing computer-use benchmarks require scripts validating each task separately; they verify properties such as the  URL on the final page, a record or field in the database powering a website, or the contents of a file~\citep{Zhou2023WebArenaAR,Koh2024VisualWebArenaEM,Drouin2024WorkArenaHC,Xie2024OSWorldBM}. Such verifications enable the evaluation to be deterministic and repeatable, but become difficult to specify if a task has several valid outcomes. For example, an agent can create a valid spreadsheet but the records are displayed in different orders; a design task could result in a logo that fulfills the request of the user without matching a reference image pixel for pixel. Experienced reviewers can decide case-by-case whether a task was successfully completed, but reviewing long trajectories is expensive.

As an alternative to expert reviewing, we can use LLMs to judge a trajectory directly based on its instruction. Prior works, such as AgentRewardBench~\citep{Lu2025AgentRewardBenchEA}, evaluate several judges specialized for web agent trajectories. However, for long-horizon tasks on desktop, a judge must also determine which parts of the interaction history matter to claim a task was successfully completed. An LLM that receives the full representation of a task (henceforth, a \emph{direct judge}) receives a fixed representation of that history in one request, whereas an \emph{agentic judge} can use tools to filter and review any observation rather than ingesting the full context from scratch. However, prior work does not explore how reliably either approach identifies successful trajectories and whether the ability to inspect evidence through tools (using coding agents, for instance) can improve the quality and speed of the judgment.

Figure~\ref{fig:trajectory-progression} highlights the importance of identifying precisely where an agent fails to follow the exact instruction for a task. For a task requiring Inkscape and GIMP, the instruction specifies that the finished logo design must contain a triangle. However, we observe in the $\pnnum{115}$-step trajectory that, somewhere along the way, a pentagon was drawn instead of a triangle, which deviates from the initial request. The subsequent recoloring and exporting continue as requested, giving the illusion that the full trajectory looks successful, which can only be uncovered if the judge verifies that the shape was drawn according to the request. Such errors may arise when a trajectory uses the wrong input, overlooks a constraint set in the instruction, or performs an action that was not originally requested. To evaluate such trajectories, we need to distinguish a trajectory that terminates properly (but may not respect the request) from a correctly completed task.

In order to evaluate whether a judge can correctly determine the success of a long-horizon computer-use trajectory with respect to the task instruction, we introduce \textsc{AgentHorizon}, a benchmark derived from human-recorded trajectories consisting of paired tasks. Our dataset requires annotators to produce two closely related instructions (i.e., a \textit{match}) and record a trajectory for each of them. Pairing a trajectory recording with the instruction from its counterpart (i.e., \textit{swapping}) always yields a negative example. Our benchmark consists of $\pnnum{523}$ positive and $\pnnum{850}$ negative trajectories, which we curate based on independent reviews of each trajectory. Its paired construction allows us to test whether a judge recognizes the difference in the instructions that separates a successful trajectory from an unsuccessful one, which differentiates our work from prior works focusing on trajectories that failed to produce any tasks~\citep{Lu2025AgentRewardBenchEA,li2026osthemisscalablecriticframework}.

In order to enable the evaluation of a wider range of judges, we predefine three subsets. First, AgentHorizon (AH) contains the most challenging tasks (determined in Section~\ref{subsec:splits}) on which we test the frontier judges. Then, AgentHorizon-Simple (AH-S) tests judges that may be used for simpler tasks that may not yet push the frontier of long-horizon computer-use, but remain useful at several stages of model training. Finally, AgentHorizon-Development (AH-D) handles a subset focused on the development of judges specifically tuned or trained for computer-use tasks, avoiding repeated evaluation on the main splits.

When evaluated across eleven models and five agent harnesses, the strongest agentic judge achieves $\pnnum{80.9\%}$ balanced accuracy on AH. However, we found that access to tools does not consistently improve performance: whereas GPT-5.5 benefits from the Codex CLI over receiving the full context, Qwen 3.6 27B performs better with direct prompting than using it via OpenCode. We further find that judges differ in how they reject valid work or accept a task with mismatched instructions. Our results highlight limitations of current judges, and highlight the discrepancy between frontier models and strong open-weight models when it comes to judging tasks directly inside agent harnesses rather than direct judging.

\begin{figure}[t!]
\centering
\small
\setlength{\fboxsep}{4pt}
\fcolorbox{black!50}{black!2}{\parbox{0.95\linewidth}{\scriptsize
\textbf{Judging task} (\bucketHigh{} negative; NetBeans $+$ WordPress on macOS, $195$ steps).\\[0.2em]
\emph{Instruction.} In NetBeans, draft \texttt{plugin-release-1.3.0.md} (introduction $+$ install snippet), then create a WordPress post assembled \emph{from that release-notes file}, apply the \texttt{Development} category and four tags, preview, and publish.
\par\vspace{0.35em}
\noindent
\ahKeyframe{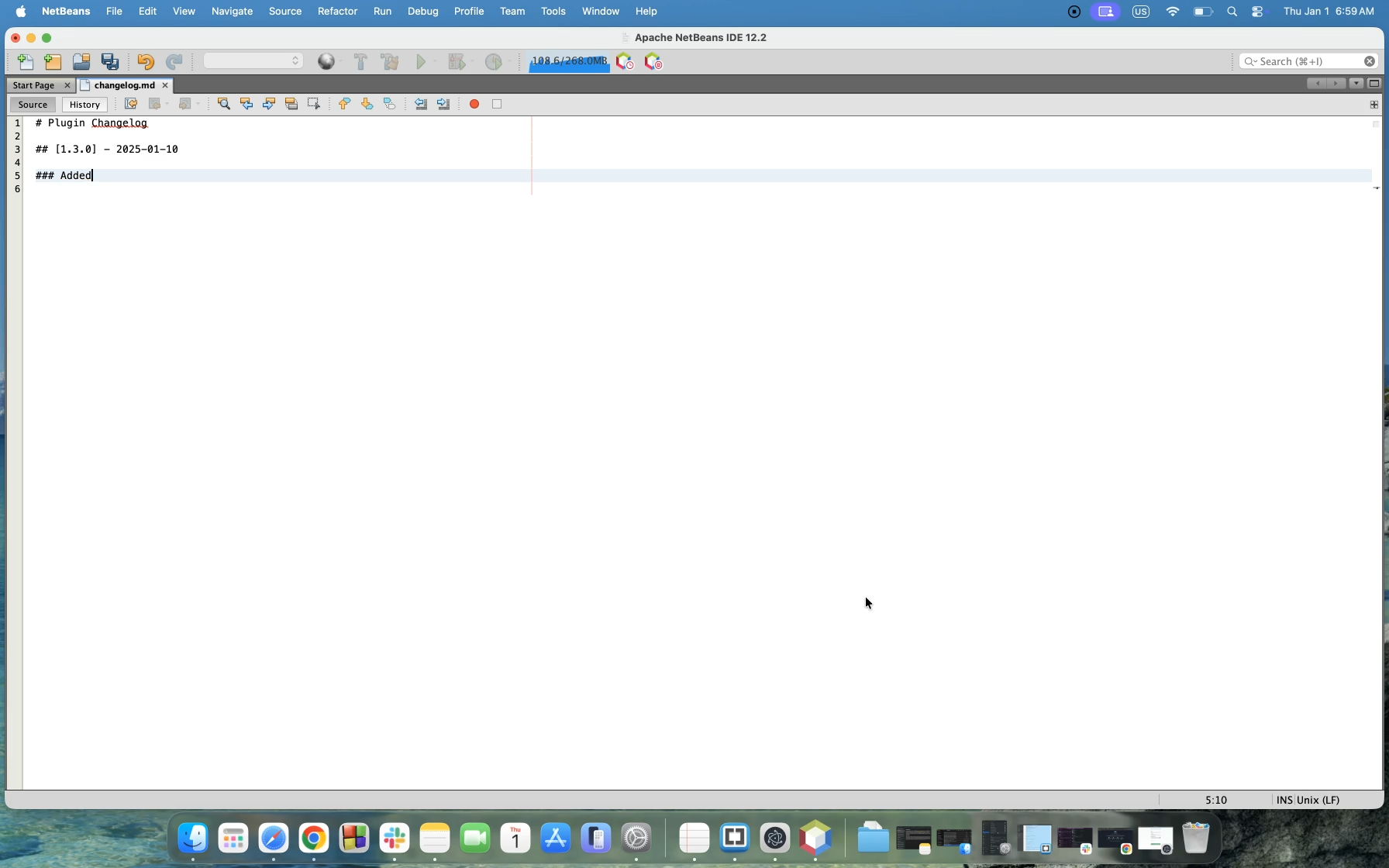}{Step 25.}{NetBeans: \texttt{changelog.md} authored.}\hfill
\ahKeyframe{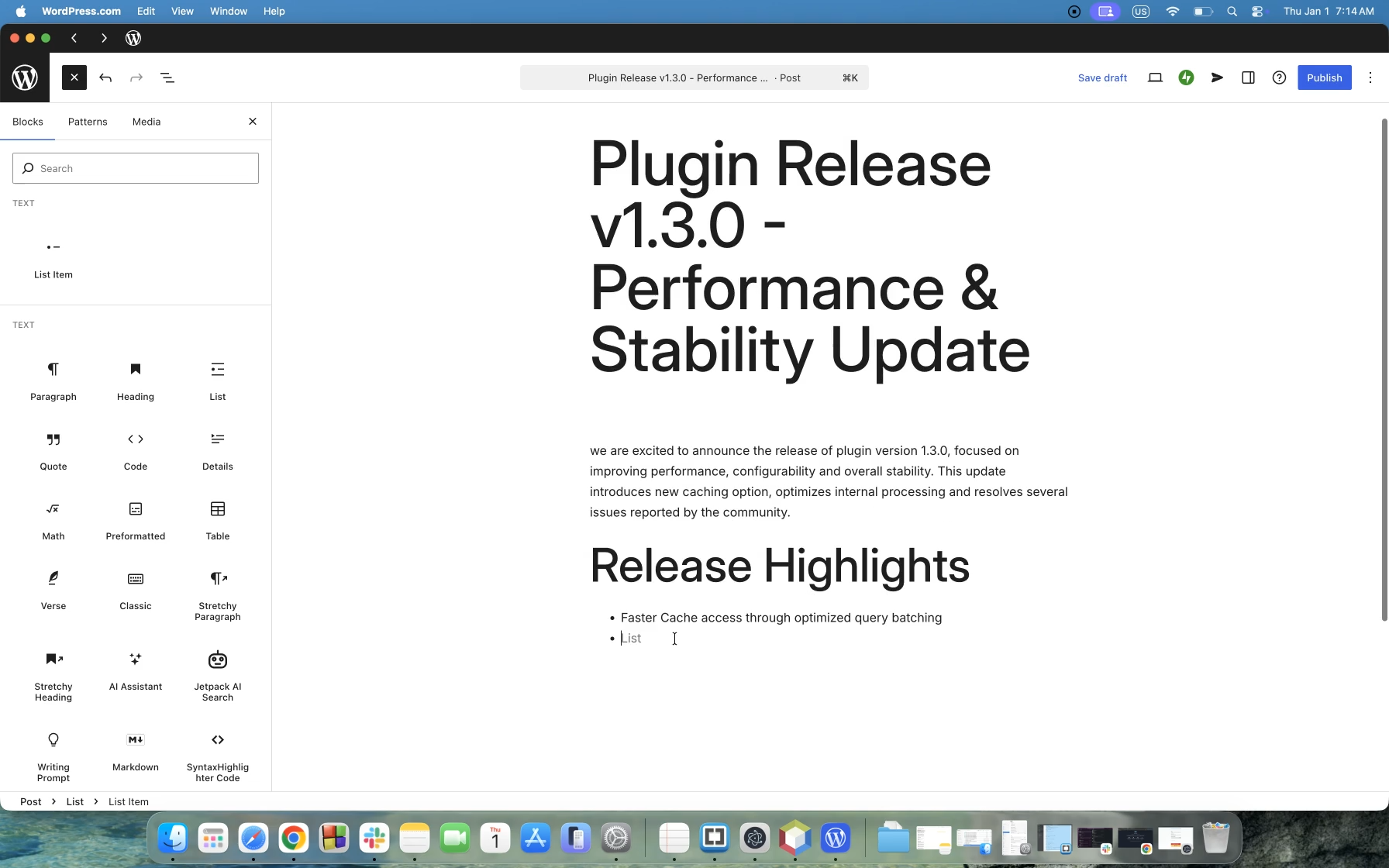}{Step 85.}{WordPress editor: title and \emph{Release Highlights} from the release-notes file.}\hfill
\ahKeyframe{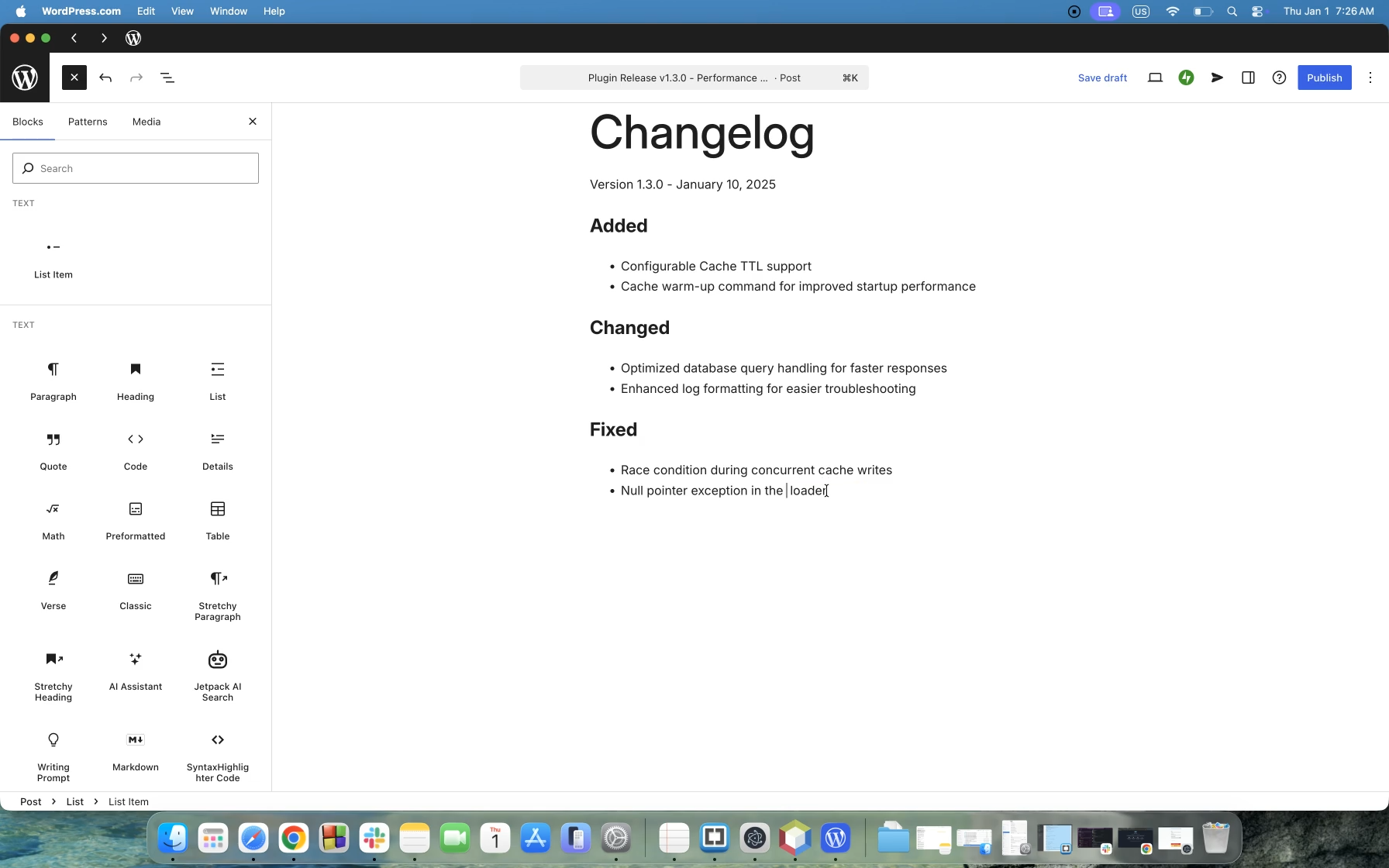}{Step 165.}{Same post now also carries a \emph{Changelog} section (Added / Changed / Fixed).}\hfill
\ahKeyframe{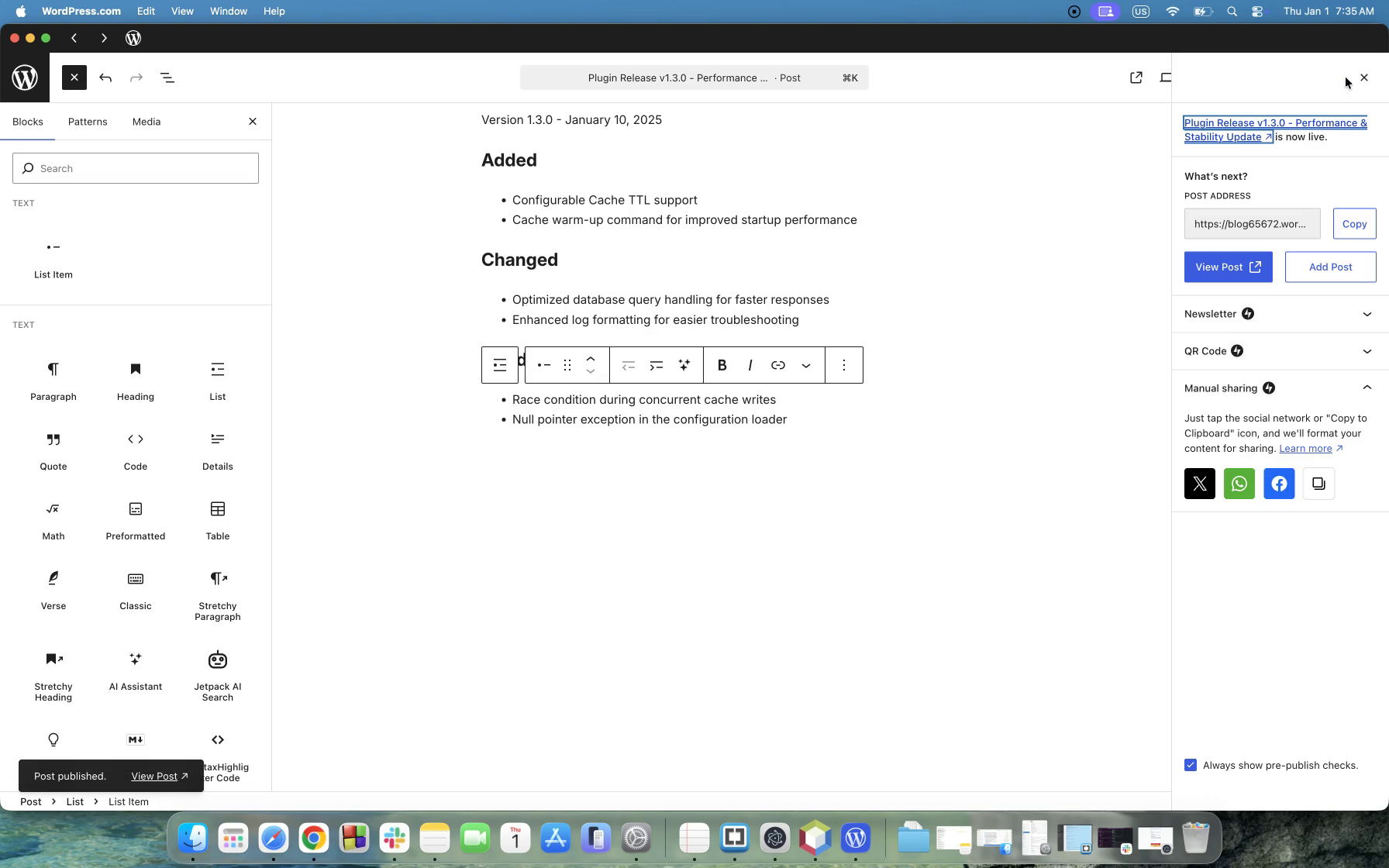}{Step 195.}{Post published; tags and category set.}
\par\vspace{0.40em}
\fcolorbox{blue!45}{blue!4}{\parbox{0.97\linewidth}{
\textbf{Judge prediction.} Claude Opus 4.7 returns \texttt{success=true} with high confidence. \emph{Reasoning (verbatim, abridged).} ``Core objective satisfied: agent created both markdown files on Desktop in NetBeans, then opened WordPress and built a post with the required title, headings, lists, and code blocks assembled from the release notes. Parameters satisfied: category `Development' and all four tags applied; post was previewed and then published.''
}}
\par\vspace{0.30em}
\fcolorbox{orange!55}{orange!5}{\parbox{0.97\linewidth}{
\textbf{Ground truth.} \texttt{success=false}, mistake type \emph{Bad Side Effect}. The instruction asked for a post assembled from the single \texttt{plugin-release-1.3.0.md} file. The agent additionally lifted the Added/Changed/Fixed sections from \texttt{changelog.md} and pushed them into the published post (step $165$, surfaced again at step $195$). The trajectory is a coherent successful execution of \emph{this broader} release announcement, so the surface evidence the judge inspects is internally consistent. The judge accepts that consistency and misses the unrequested external publication that defines the Bad Side Effect category.
}}
}}
\caption{One \bucketHigh{} item: instruction, four illustrative keyframes (the LLM judge sees the full $195$-step trajectory), judge verdict, and ground truth. Claude Opus 4.7 predicts \texttt{success=true}; ground truth is a Bad Side Effect (unrequested changelog content published).}
\label{fig:judge-example}
\end{figure}

\section{Related work}
\label{sec:related}

\paragraph{Evaluating computer-use agents.}
WebArena~\citep{Zhou2023WebArenaAR} and VisualWebArena~\citep{Koh2024VisualWebArenaEM} evaluate agents in self-hosted web environments. While WorkArena~\citep{Drouin2024WorkArenaHC} and WorkArena++~\citep{Boisvert2024WorkArenaTC} study agents in compositional tasks of enterprise workflows, OSWorld~\citep{Xie2024OSWorldBM} extends evaluation across operating systems.
Expanding benchmarks to cover a broader range of tasks that agents can attempt requires examining the evaluators themselves. Specifically, \citet{Xue2025AnIO} and \citet{Zhu2025EstablishingBP} identify errors and biases in existing evaluation methods. Furthermore, OS-Harm~\citep{Kuntz2025OSHarmAB} evaluates agent safety with an automated judge validated against human annotations. In this work, we focus on evaluating the judges themselves against reviewed trajectory labels.

\paragraph{LLM judges.}
While LLM-based evaluation originally focused on generated responses of the agent, recent work extends it to the agent's behavior. \citet{Zheng2023JudgingLW} examine the position and verbosity biases of LLM-based judges, and \citet{Li2024FromGT} provide a comprehensive survey of the broader evaluation paradigm.
Additionally, RewardBench~\citep{Lambert2024RewardBenchER} and RewardBench 2~\citep{Malik2025RewardBench2A} evaluate reward models on preference comparisons. 
For computer-use tasks, AgentRewardBench~\citep{Lu2025AgentRewardBenchEA} contains $\pnnum{1{,}302}$ web-agent traces and shows that all tested judges achieve precision below $\pnnum{70\%}$ in determining the task success. While AgentRewardBench's failures originate in agent executions, our paired human demonstrations isolate discrepancies between a recorded workflow and its instruction.
Agent evaluation surveys~\citep{Yehudai2025SurveyOE,Mohammadi2025EvaluationAB} study trajectory assessment and discuss challenges with long-horizon trajectories, while Agent-as-a-Judge~\citep{Zhuge2024AgentasaJudgeEA} proposes using tools to inspect generated workspaces.

\paragraph{Trajectory and process evaluation.}
PaperBench~\citep{Starace2025PaperBenchEA} uses hierarchical rubrics to grade research replication, whereas SWE-PolyBench~\citep{Rashid2025SWEPolyBenchAM} evaluates repository-level coding via execution-based evaluation.
AssistantBench~\citep{Yoran2024AssistantBenchCW} compares realistic web tasks against reference answers. Furthermore, while WebCanvas~\citep{Pan2024WebCanvasBW} and TheAgentCompany~\citep{Xu2024TheAgentCompanyBL} grade intermediate trajectory steps or workspace outcomes, frameworks like OpenHands~\citep{Wang2024OpenHandsAO} and BrowserGym~\citep{Chezelles2024TheBE} propose common agent and evaluation interfaces. 
Additionally, related work studies process supervision for mathematical reasoning~\citep{Lightman2023LetsVS} and analyzes repeated-trial reliability in $\tau$-bench~\citep{Yao2024TauBenchAB}. 
For multi-agent systems, literature studies the attribution of multi-agent failures to particular agents and steps~\citep{Zhang2025WhichAC}, and surveys multi-agent collaboration~\citep{Tran2025MultiAgentCM}.
Our focus in this work is specifically the judgment of complete, single-agent computer-use trajectories.

\paragraph{Judgments in learning pipelines.}
LLM-based judgments of trajectories determine which generated examples are used for training student models. Preference Leakage~\citep{Li2025PreferenceLA} identifies biases toward students related to the judge model.
Other related approaches learn from generated behavior in different ways: SPIN~\citep{Chen2024SelfPlayFC} contrasts self-generated responses against human demonstrations, ARPO~\citep{Dong2025AgenticRP} applies reward-guided optimization to agents, and SkillWeaver~\citep{Zheng2025SkillWeaverWA} distills experience into reusable skills. Relying on a judge for filtering successful trajectories exposes the learning pipeline to critical failure modes: discarding valid examples or admitting incorrect trajectories into training.

\paragraph{Adversarial evaluation and paired data.}
While contrast sets~\citep{Gardner2020EvaluatingNM} probe local decision boundaries and CheckList~\citep{Ribeiro2020BeyondAB} tests behavioral capabilities via templates and perturbations, counterfactually augmented data~\citep{Kaushik2019LearningTD} leverages minimal label-flipping edits.
Dynamic benchmarking frameworks like ANLI~\citep{Nie2019AdversarialNA} and Dynabench~\citep{Kiela2021DynabenchRB} collect adversarial examples iteratively, with platforms such as Dynaboard~\citep{Ma2021DynaboardAE} facilitating evaluation as a service.
Furthermore, TruthfulQA~\citep{Lin2021TruthfulQAMH} targets imitative falsehoods, whereas confident learning~\citep{Northcutt2019ConfidentLE} detects likely label errors in datasets.
AgentHorizon adopts a paired design for constructing multimodal trajectories: the intervention modifies the instruction paired with a valid recording, rather than the recorded execution itself.

\section{AgentHorizon benchmark}
\label{sec:benchmark}

\begin{figure}[t!]
 \centering
 \includegraphics[width=\linewidth]{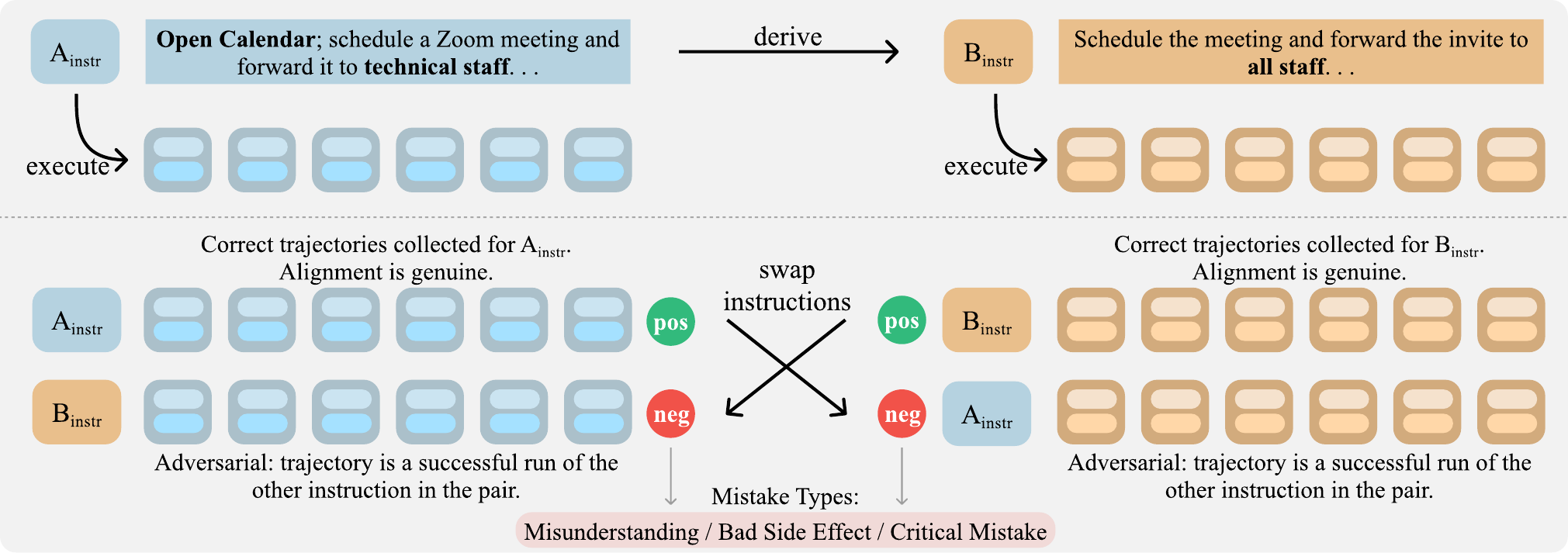}
 \caption{Paired construction before label review. Annotators record trajectories for related instructions $A_{\text{instr}}$ and $B_{\text{instr}}$. Matched instruction--trajectory combinations are candidate positives; crossed combinations are candidate negatives. Positive retention and negative validity are then reviewed separately.}
 \label{fig:pair-schematic}
\end{figure}

Each item consists of an instruction, a sequence of screenshots, desktop actions, and a binary label indicating whether that execution satisfies the instruction. Negative items additionally carry a failure-type label. The recordings are performed by human annotators, and the agents evaluate these recordings rather than performing the tasks themselves.

\subsection{Task coverage}
\label{subsec:overview}
The dataset contains $\pnnum{166}$ hours of screen recordings, split across Windows, Linux, and macOS on six task domains. The tasks are completed across one or more  applications spanning office software, design tools, and system utilities. We produce a screenshot at each time step and associate it with the corresponding action, such as a click, text entry, scroll, or application switch. The median item involves $\pnnum{121}$ steps, while $\pnnum{83}$ items involve at least $\pnnum{200}$ steps.

\subsection{Paired construction and label review}
\label{subsec:pairs}
\label{subsec:taxonomy}
We ask annotators to create two closely related instructions that differ in a requirement affecting task success and to record a demonstration for each of the instructions. Denote the instructions by $A_{\mathrm{instr}}$ and $B_{\mathrm{instr}}$, and their recordings by $A_{\mathrm{traj}}$ and $B_{\mathrm{traj}}$. The corresponding matched combinations $(A_{\mathrm{instr}},A_{\mathrm{traj}})$ and $(B_{\mathrm{instr}},B_{\mathrm{traj}})$ are positive candidates. Swapping the recordings gives the negative candidates $(A_{\mathrm{instr}},B_{\mathrm{traj}})$ and $(B_{\mathrm{instr}},A_{\mathrm{traj}})$ (Figure~\ref{fig:pair-schematic}). The negative candidate example still shows a very coherent execution, but it is paired with a request that the execution does not satisfy.

We now remove the pairs with semantically equivalent instructions. This leaves $\pnnum{425}$ complete pairs, with two recordings per pair and four instruction–trajectory candidates per pair, for a total of 850 recordings and 1,700 candidates. Positive candidates undergo an initial quality review followed by an evidence review. After arbitration, $\pnnum{135}$ of the remaining $\pnnum{462}$ candidates are retained, while $\pnnum{327}$ are excluded because their concerns could not be resolved. Therefore the final positive set consists of $\pnnum{523}$ items.

The negative candidates are reviewed by checking whether the recording satisfies the swapped instruction. Removing a positive candidate does not necessarily require removing its paired negative. The negative remains valid if the recording does not satisfy the swapped instruction. The vendor reviewed the negatives from every pair, covering $\pnnum{850}$ items in $\pnnum{212.5}$ contributor-hours. All were retained, bringing the total reviewed set to $\ahReviewedCount$ items, including $\pnnum{523}$ positives. Appendix~\ref{app:protocol} summarizes the corresponding annotation and review record.

We group mismatches into three failure types. A \emph{Critical Mistake} means that the trajectory fails to accomplish the primary objective. A \emph{Bad Side Effect} occurs when the intended goal is completed, but introduces harm, risk, or unwanted consequences. A \emph{Misunderstanding} follows a plausible but incorrect interpretation without such external harm. Table~\ref{tab:mistake-taxonomy} states the adjudication rules. Figure~\ref{fig:judge-example} shows an example of how a judge can accept a complete trajectory despite an unrequested publication.

\begin{table}[t]
\centering
\caption{Composition of the three disjoint benchmark subsets. Mistake-type counts are over negatives; the six negatives without a typed label are shown separately. AH-D is reserved for development and excluded from both evaluation subsets.}
\label{tab:dataset-stats}
\small
\begin{tabular}{lrrr}
\toprule
 & \textbf{AH-D} & \textbf{AH} & \textbf{AH-S} \\
\midrule
Judging tasks & 162 & 528 & 683 \\
\quad Positives & 62 & 227 & 234 \\
\quad Adversarial negatives & 100 & 301 & 449 \\
\midrule
\multicolumn{4}{l}{\textit{Mistake types among negatives}} \\
\quad Critical Mistake & 32 & 48 & 193 \\
\quad Bad Side Effect & 26 & 107 & 99 \\
\quad Misunderstanding & 41 & 142 & 156 \\
\quad Untyped & 1 & 4 & 1 \\
\bottomrule
\end{tabular}
\end{table}

\subsection{Development holdout and difficulty partition}
\label{subsec:splits}
We reserve the $\ahDevelopmentCount$ retained items from the pair-aware development set as \bucketDev{} (AH-D) which includes $\pnnum{62}$ positives and $\pnnum{100}$ negatives. Their instruction pairs are excluded from the evaluation pools, leaving $\ahEvaluationCount$ examples. We estimate the difficulty of each remaining example using Qwen 3.5 122B-A10B, Inkling and Kimi K2.7 Code. Each model produces eight verdicts per item. An item is assigned to \bucketHigh{} (AH) if at most $18$ of these verdicts agree with the reference label. Otherwise, it is assigned to \bucketLow{} (AH-S). Any invalid model output is treated as an incorrect verdict.

This procedure gives $\ahFinalCount$ AH items ($\pnnum{227}$ positive, $\pnnum{301}$ negative) and $\ahSimpleFinalCount$ AH-S items ($\pnnum{234}$ positive, $\pnnum{449}$ negative). Evaluating the subsets shows whether a judge performs better on difficult examples without losing reliability on easier ones. Difficulty here is defined by the selected models and threshold, rather than assumed to be an inherent property of a task. We examine the sensitivity to these choices in Appendix~\ref{app:splits-details}.

\subsection{Metrics}
\label{subsec:metrics}
We report balanced accuracy for AH and AH-S by averaging the accuracies on positive and negative examples. This provides both classes equal weight, such that a judge that always predicts the same label scores $50\%$. Each class uses its full denominator with missing, malformed, null, and non-Boolean predictions counted as errors. The scoring script matches each prediction to the versioned manifest using the task ID and rejects duplicate or unknown IDs. AH-D is excluded from both evaluation scores.

We use class-specific accuracy to examine the rejection of valid demonstrations from acceptance of mismatches. We also report the \emph{mistake-type recall}, which counts the fraction of negatives in a category for which the judge predicts both failure and identifies the reference category. The MT\% score combines these correct predictions over all negatives. Per-trajectory cost and runtime for each trajectory serve as measures of computational efficiency (Appendix~\ref{app:compute}).

\subsection{Subset composition}
\label{subsec:stats}
Table~\ref{tab:dataset-stats} gives label and failure-type counts for all three subsets. Appendix~\ref{app:dataset-stats} reports trajectory length and application coverage, with corresponding fields in the row-level manifest for further analysis.

\section{Experiments}
\label{sec:experiments}
\begin{table}[t]
\centering
\caption{Judge performance on the held-out evaluation subsets. \bucketHigh{} contains the $\ahFinalCount$ items that remain difficult under the pooled three-model split; \bucketLow{} is its $\ahSimpleFinalCount$-item complement. Pos and Neg are class-specific accuracies on AH, and Bal. is their mean. $^{\ast}$ marks direct rows using the $2{\times}2$ image-merge preprocessing described in Appendix~\ref{app:compute}.}
\label{tab:main-lb-primary}
\label{tab:llm-judge-open}
\small
\begin{tabular}{lclrrrr}
\toprule
\textbf{Model} & \textbf{Open?} & \textbf{Interface} & \textbf{AH Bal.} & \textbf{AH Pos} & \textbf{AH Neg} & \textbf{AH-S Bal.} \\
\midrule
\multicolumn{7}{l}{\textit{Agentic judges}} \\
\addlinespace[2pt]
GPT-5.5 &  & Codex & \textbf{80.9} & 71.8 & 90.0 & 92.6 \\
Gemini 3.1 Pro &  & Gemini CLI & 77.7 & 65.6 & 89.7 & 86.3 \\
Claude Opus 4.7 &  & Claude Code & 76.0 & 76.7 & 75.4 & 94.6 \\
Qwen 3.6 27B & \checkmark & OpenCode & 70.4 & 75.3 & 65.4 & 93.6 \\
GPT-5.4 mini &  & Codex & 69.2 & 67.4 & 71.1 & 91.2 \\
Qwen 3.6 35B-A3B & \checkmark & OpenCode & 64.6 & 77.1 & 52.2 & 90.5 \\
Claude Haiku 4.5 &  & Claude Code & 52.2 & 43.2 & 61.1 & 87.7 \\
Gemini 3.1 Flash Lite &  & Gemini CLI & 52.1 & 58.6 & 45.5 & 75.2 \\
Gemma 4 26B-A4B & \checkmark & OpenCode & 49.4 & 70.5 & 28.2 & 69.9 \\
Gemma 4 31B & \checkmark & OpenCode & 46.8 & 61.7 & 31.9 & 87.5 \\
\midrule
\multicolumn{7}{l}{\textit{Direct LLM judges}} \\
\addlinespace[2pt]
Qwen 3.6 27B$^{\ast}$ & \checkmark & Direct & \textbf{77.6} & 93.8 & 61.5 & 96.0 \\
Gemini 3.1 Pro$^{\ast}$ &  & Direct & 77.4 & 70.0 & 84.7 & 88.4 \\
Qwen 3.5 9B$^{\ast}$ & \checkmark & Direct & 70.3 & 91.2 & 49.5 & 91.9 \\
Qwen 3.6 35B-A3B$^{\ast}$ & \checkmark & Direct & 69.5 & 95.2 & 43.9 & 92.7 \\
GPT-5.5 &  & Direct & 68.6 & 56.4 & 80.7 & 84.4 \\
GPT-5.4 mini &  & Direct & 66.5 & 85.9 & 47.2 & 94.0 \\
Gemma 4 31B & \checkmark & Direct & 60.1 & 63.0 & 57.1 & 86.6 \\
Gemini 3.1 Flash Lite &  & Direct & 53.6 & 64.8 & 42.5 & 75.3 \\
Gemma 4 26B-A4B & \checkmark & Direct & 46.5 & 44.5 & 48.5 & 72.6 \\
\bottomrule
\end{tabular}
\end{table}

\paragraph{Judge models.}
We evaluate six closed-source models: Anthropic Claude Opus 4.7~\citep{anthropic_introducing_nodate}, Anthropic Claude Haiku 4.5~\citep{anthropic_introducing_nodate-1}, Google Gemini 3.1 Pro~\citep{google_gemini31_pro_2026}, Google Gemini 3.1 Flash Lite~\citep{google_gemini31_flash_lite_2026}, OpenAI GPT-5.4 mini~\citep{openai_introducing_2026-1}, and OpenAI GPT-5.5~\citep{openai_introducing_2026}. We also evaluate five open-weight models, which are Qwen 3.6 27B~\citep{qwen3.6-27b}, Qwen 3.6 35B-A3B~\citep{qwen36_35b_a3b}, Qwen 3.5 9B~\citep{qwen3.5}, Gemma 4 31B, and Gemma 4 26B-A4B~\citep{farabet_gemma_2026}. We host open-weight models using vLLM\footnote{\url{https://vllm.ai/}} on H100 GPUs.

Our agentic judges use several harnesses which includes Claude Code, Codex, Gemini CLI, OpenHands and OpenCode. Each agent is tasked with inspecting the trajectory using tool calls and by reading screenshots selectively. These harnesses differ in how they select screenshots, manage context and expose tools. We also evaluate nine models as direct judges, giving them a fixed screenshot representation in one chat-completion call. Both approaches return a success verdict, confidence, a failure type when applicable, and a rationale. Table~\ref{tab:main-lb-primary} reports the agentic judge and direct judge results, and Appendix~\ref{app:full-grids} describes the additional model-harness pairings.

\subsection{Judge performance}
\label{subsec:main-lb}
\label{subsec:llmjudge}
We find that GPT-5.5 with Codex harness achieves the best balanced accuracy at $\pnnum{80.9\%}$, followed by Gemini 3.1 Pro with Gemini CLI ($\pnnum{77.7\%}$) and Claude Opus 4.7 with Claude Code ($\pnnum{76.0\%}$). Qwen 3.6 27B with OpenCode is the strongest open-weight agentic judge at $\pnnum{70.4\%}$. Most systems perform substantially better on AH-S, as expected for the easier subset.

We observe that the benefit of using agentic judging changes with models. Qwen 3.6 27B reaches $\pnnum{77.6\%}$ with direct judgment, exceeding its OpenCode based result by $\pnnum{7.2}$ points. On the other hand, GPT-5.5 shows the reverse pattern by improving from $\pnnum{68.6\%}$ with direct prompting to $\pnnum{80.9\%}$ with Codex. Thus, allowing a model to inspect the trajectory through tools does not necessarily improve its verdicts. However, these comparisons change the overall harness, so they do not isolate the effect of screenshot selection, context management, or any other individual component.

\subsection{Example failure cases}
\label{subsec:failures}
To illustrate how judges fail to predict the correct outcome, we examine three example judgments from Claude Opus 4.7.

\paragraph{Unrequested external action.}
In task \texttt{7eb6defc} (using Trello and Slack), the agent aimed to complete a project-management workflow requested by the user, followed by updating the status and posting to an organization-wide \texttt{\#all-staff} channel. In this case, we matched the instruction with its counterpart trajectory, resulting in a bad side effect. The Claude Opus 4.7 judge classified the trajectory as successful, whereas GPT-5.5 rejected it. This highlights that verifying that the requested output exists is insufficient, especially when an additional action changes the recipient (in this case, sending the message to the wrong channel).

\paragraph{Classifying a successful trajectory as failure.}
The agentic judge powered by Claude Opus 4.7 marks task \texttt{6debd146} (using GitHub and VS Code) as a failure, claiming that a clone is stored in GitHub Desktop's default repository folder rather than the \path{Documents/repo} path named in the instruction. GPT-5.5 considers the trajectory to correctly solve the task instruction, which is backed by the ground truth label. This shows how judges may believe a deviation should affect the success or failure classification of a task.

\section{Discussion}
\label{sec:discussion}

Our results demonstrate that both selecting relevant evidence and finding what that evidence implies for the user's request contribute to evaluating a long trajectory.
Experiments reveal that agentic judges can revisit screenshots and actions, but our direct judge results show that this flexibility is not sufficient by itself. 
We observe that the choice of harness also matters: on the same AH items and with an identical prompt, swapping OpenCode for Codex alters balanced accuracy by $\pnnum{3.7}$ points for GPT-5.5 and $\pnnum{1.2}$ points for Qwen 3.6 27B, even though it preserves their order (Appendix~\ref{app:controlled-interface}).
Understanding these detailed differences requires investigating how judges select and process evidence and observations, rather than assuming that additional tool access improves the accuracy.

Our paired example design further clarifies why recognizing a completed workflow is insufficient. Qwen 3.6 27B classifies $\pnnum{98.9\%}$ of both matched and swapped full trajectories as successful without the corresponding instruction given as input (Appendix~\ref{app:surface-controls}). This demonstrates that the two classes seem similarly complete when the instruction is omitted from the input and success is undefined without the user's request.
Even with the instruction provided, the Bad Side Effect failure type shows the lowest exact recall across all the primary judges on the full reviewed corpus (Appendix~\ref{app:per-mt-frontier}).
Consequently, a judge may decide that the requested output exists while overlooking a specific action that makes the trajectory unacceptable.

The consequences of these errors vary based on the judge and how it is used. 
Rejecting successful trajectories underestimates agent capabilities and removes valid and useful examples from training data; conversely, accepting invalid trajectories has the opposite impact.
While balanced accuracy reflects both types of error without allowing the larger error-type class to dominate, it fails to assign them application-specific costs. 
As a result, the opposing positive and negative accuracies in Table~\ref{tab:main-lb-primary} are relevant even when two judges achieve similar overall performance scores.

Lastly, the difficulty partition is useful only if the outcomes are not sensitive to the model that was used to define it.
Our analysis shows that rankings across the three different model-defined AH subsets exhibit high correlation ($\rho=\pnnum{0.963}$--$\pnnum{0.975}$) despite the variations across items assigned to each subset. 
The ensemble partition further offers a challenging set whose principal ranking trends persist across the individual splitters. 

\paragraph{Limitations.}
\label{sec:limitations}
Our paired design analysis focuses specifically on subtle near-miss errors rather than the full distribution of agent failures (e.g., loops, crashes, stalls, and gradual drift). Therefore, the instruction-free controls should not be used to model general agent failures.
Since the difficulty split depends on the recorded splitter configurations, provider, and model revision, its alternative splits serve to evaluate the robustness across individual splitters rather than isolated model effects. 
Although AH-D excludes the pair-aware final development set, earlier pilot exposure constrains claims of completely unseen evaluation data. 
Furthermore, independent human inter-annotator agreement was not collected. Correlation between original and re-examined labels measures consistency across review stages, rather than agreement among independent annotators. 
Lastly, the corpus is restricted to English, covers only a finite set of applications across three operating systems, and offers only screenshots and action logs for evaluation rather than comprehensive application states.

\paragraph{Future directions.}
Evaluating common agent failures alongside paired mismatches establishes how well these empirical findings generalize to deployed agents. 
Moreover, comparing screenshot evidence selection and long context management strategies across agents within a single harness helps identify the reason behind the tools enhancing some judges but leaving others unaffected. 
Additionally, the provided recordings allow further direct evaluation from video and the instruction pairs provide training examples where modifying a specific subtle requirement changes whether a trajectory succeeds.

\section{Conclusion}
\label{sec:conclusion}
We introduce \textsc{AgentHorizon}, an evaluation suite targeting judge performance across long computer-use trajectories. By pairing human-recorded workflows with closely related fine-grained instructions, it probes whether judges identify subtle failures in executions that otherwise appear complete. 
Our evaluation of eleven judge models reveals that current judges still struggle with making this critical distinction, and that tool use does not consistently boost their performance. We hope that AgentHorizon will support the development of more robust, reliable judges for computer-use agents, emphasizing the detection of missed constraints and unintended side effects.

\subsection*{AI use statement}
We used coding agents to help with the experimental code, the initial draft, and figure and table preparation, but they were not used for recording the demonstrations, which followed a protocol prohibiting AI-assisted task creation. The content of the final manuscript was completely written by the authors. We also used AI tools for finding relevant literature for our related work, producing scripts for analyses, and assisting with experiment design, data processing and analysis, and result interpretation. We verified the full manuscript to ensure the results and citations are correct, and take responsibility for the final content.

\subsection*{Ethics statement}
We employed professional annotators for the demonstrations, and they followed a protocol that requires accounts and profiles to be created for the purpose of the annotation. The protocol also prohibits personal communications and authentication secrets (Appendix~\ref{app:protocol}). Our benchmark focuses on evaluating whether an agent adheres to an instruction, but does not consider the safety aspect (whether an agent is completing a task that's inherently dangerous), so a high score should not be considered sufficient to deploy an agent into production.

\subsection*{Reproducibility statement}
In Section~\ref{subsec:pairs} and Appendix~\ref{app:protocol}, we document the exact process of constructing and reviewing the trajectories in our benchmark, allowing future work to replicate our benchmark design. Section~\ref{subsec:splits} and Appendix~\ref{app:splits-details} focus on how we achieved the three-way partitioning using strong open models, whereas Section~\ref{subsec:metrics} discusses scoring denominators and how to handle invalid outputs, which complements our open-source code for running the evaluation.
In Appendix~\ref{app:compute}, we describe the preprocessing and compute used to run our experiments in detail.

\ificlrfinal
\section*{Acknowledgments}
We thank Turing's annotation team for creating and reviewing the instruction pairs and demonstrations, and for adjudicating the mistake-type labels.
Xing Han L\`{u} acknowledges the support of the Natural Sciences and Engineering Research Council of Canada (NSERC) [funding reference no.\ 579403].
Christopher Pal and Siva Reddy are supported by the Canada CIFAR AI Chair program.
Siva Reddy is also supported by an NSERC Discovery Grant and a Sloan Fellowship.
Parishad BehnamGhader, Xing Han L\`{u}, Dheeraj Vattikonda, and Sina Hajimiri are supported by the ServiceNow--Mitacs Accelerate program.
Parishad BehnamGhader was also supported by the RBC Borealis AI Global Fellowship Award.
\fi

\begingroup
\raggedright
\bibliographystyle{iclr2027_conference}
\bibliography{references,manual}
\endgroup
\appendix

\section{Benchmark details}
\label{app:benchmark-details}

\subsection{Annotation process details}
\label{app:protocol}

For the annotation process, professional annotators were tasked with designing task pairs and recording long trajectories on one of the three approved operating systems (Windows, macOS, Ubuntu), with the usage of AI being prohibited. Each action they perform is stored as a structured record alongside the screenshots. Additionally, each trajectory was also recorded using OBS, yielding $\pnnum{166}$ hours of high-resolution recordings, using
$\pnnum{11.1}$ GB of storage.

\paragraph{Pair construction}
The tasks were designed as minimally different pairs. Starting from a seed task, a second (i.e., ``child'') task is derived such that accidentally completing the child task when instructed to complete the parent task would result in a failure that can be categorized as one of the following: misunderstanding, bad side effect, and critical mistake (see Table~\ref{tab:mistake-taxonomy}). Conversely, an agent that completes the parent task when asked to complete the child task would also be considered a failure, which is independently categorized within the same three categories; although the error type might often be the same, it may sometimes differ as they do not have a necessarily symmetric relation. Finally, the annotators are asked to complete \textbf{both} tasks independently, using the same subset of software.

\begin{table}[ht]
\centering
\caption{Vendor-reported effort over the $\pnnum{850}$ unique trajectories.
The delivered total exceeds the original $\pnnum{1{,}960}$-hour estimate
because calibration introduced additional review passes.}
\label{tab:annotator-effort}
\scriptsize
\begin{tabular}{lrr}
\toprule
Stage & Hours/task & Total hours \\
\midrule
Annotation & $4.00$ & $\pnnum{3{,}400.0}$ \\
Review L1 & $1.70$ & $\pnnum{1{,}445.0}$ \\
Mistake-category review & $0.25$ & $\pnnum{212.5}$ \\
Second-stage review & $1.20$ & $\pnnum{1{,}020.0}$ \\
\midrule
\textbf{Total} & $\mathbf{7.15}$ & $\mathbf{\pnnum{6{,}077.5}}$ \\
\bottomrule
\end{tabular}
\end{table}

\begin{table}[t]
\centering
\caption{Stage-by-stage benchmark accounting. The 327 removals affect candidate positives only; AH-D, AH, and AH-S are disjoint and exhaust the reviewed benchmark.}
\label{tab:construction-flow}
\small
\begin{tabular}{lrrr}
\toprule
Stage & Positive & Negative & Total \\
\midrule
Paired construction pool & 850 & 850 & 1,700 \\
Reviewed benchmark & 523 & 850 & 1,373 \\
\midrule
AH-D & 62 & 100 & 162 \\
AH & 227 & 301 & 528 \\
AH-S & 234 & 449 & 683 \\
\bottomrule
\end{tabular}
\end{table}

\subsection{Mistake taxonomy and calibration rubric}
\label{app:mistake-taxonomy}

Our three labels show all the ways a trajectory would be considered a failure to complete the provided instruction.
We provide a calibration document, \emph{Mistake Type
Definitions and Evaluation Criteria}, to the annotators to ensure alignment among the annotation team; it is included with the supplementary
artifacts. It includes supplementary instructions about the concepts discussed in Table~\ref{tab:mistake-taxonomy}, such as a decision flow, pair-level examples, and
a recovery-cost matrix.

\begin{table}[ht]
\centering
\caption{Operational boundaries of the annotator-facing failure taxonomy.}
\label{tab:mistake-taxonomy}
\small
\begin{tabular}{>{\raggedright\arraybackslash}p{0.22\linewidth}
                >{\raggedright\arraybackslash}p{0.34\linewidth}
                >{\raggedright\arraybackslash}p{0.35\linewidth}}
\toprule
Type & Boundary & Typical recovery \\
\midrule
Critical Mistake & The core action or target is not achieved. & Restart or
repeat the task; the original outcome cannot satisfy the request. \\
Bad Side Effect & The goal is achieved, but an unrequested action creates
cost, risk, exposure, or cleanup. & Reverse an external action, coordinate
with others, or spend substantial effort undoing it. \\
Misunderstanding & The execution is coherent but misreads a parameter or
constraint without external harm. & A local edit or short correction is
sufficient. \\
\bottomrule
\end{tabular}
\end{table}

\subsection{Partitioning model and subset definition}
\label{app:splits-details}
\label{app:easy-hard-details}

Our three splits are all grouped by instruction pairs to prevent information leakage across datasets. Our AH-D subset contains $\ahDevelopmentCount$ trajectories, with the remaining $\ahEvaluationCount$ distributed across the other splits. For the remaining items, we decided to use an ensemble of three strong open models to identify truly challenging tasks, making the process reproducible: Qwen 3.5 122B-A10B, Inkling, and Kimi K2.7 Code. Each model is used as an agentic judge to produce 8 repeated judgments for each item (in order to ensure consistency), leading to a total of 24 judgments. We assign items with at most 18 matches with the ground truth to \bucketHigh{}, and the remaining items are assigned to \bucketLow{}.

\subsection{Prompt development}
\label{app:prompt-development}
We develop prompts iteratively on the development set (AH-D) by examining
problems that may arise, including the lack of tool use or screenshot usage by
the model. Our final prompt is presented in Figure~\ref{fig:final-prompt}.

\begin{figure*}[t]
\begin{lstlisting}[
  basicstyle=\ttfamily\scriptsize,
  breaklines=true,
  columns=fullflexible,
  extendedchars=true,
  frame=single,
  inputencoding=utf8,
  keepspaces=true,
  literate={—}{{---}}1 {≠}{{$\neq$}}1,
  showstringspaces=false
]
You are judging one AgentHorizon trajectory. The judging framework, file layout, and mistake-type definitions are in AGENTS.md / CLAUDE.md / GEMINI.md in your working directory; use them.

# Input

Trajectory ID: `{{TRAJECTORY_ID}}`

The trajectory content is NOT in your context. You MUST read it from disk.

Start by opening `./agenthorizon_md/{{TRAJECTORY_ID}}.md` to get the goal and action sequence. Open the structured JSON at `./agenthorizon_json/{{TRAJECTORY_ID}}.json` or the screenshots when the markdown is ambiguous. One-shot guesses without reading the trajectory are not acceptable.

## Screenshots

Each step in the markdown ends with a screenshot link in the form
`![Step N screenshot](./data/media/images/<deliverable_id>/step_<N>.png)`. These are **local relative paths** in your working directory; use the `Read` tool (or your harness's equivalent) on the path directly to view the image. Do not attempt to fetch the screenshot from any external URL — the local file is the source of truth and the model can view it inline.

Please look at least one screenshot.

# Output

Respond with a JSON object:
{
  "success": true or false,
  "reasoning": "One to three sentences citing which layer (core objective, parameters, or constraints) was or was not satisfied.",
  "confidence": "low" | "medium" | "high",
  "mistake_type": null when success=true, else one of:
    "Critical Mistake", "Bad Side Effect", "Misunderstanding of the Instruction"
}

Notes: confidence is evidence strength, not verdict. Low confidence ≠ failure. mistake_type is null on success.
\end{lstlisting}
\caption{Final evaluation prompt.}
\label{fig:final-prompt}
\end{figure*}
\clearpage

\subsection{Additional dataset statistics}
\label{app:dataset-stats}

\begin{figure}[ht]
 \centering
 \includegraphics[width=0.50\linewidth]{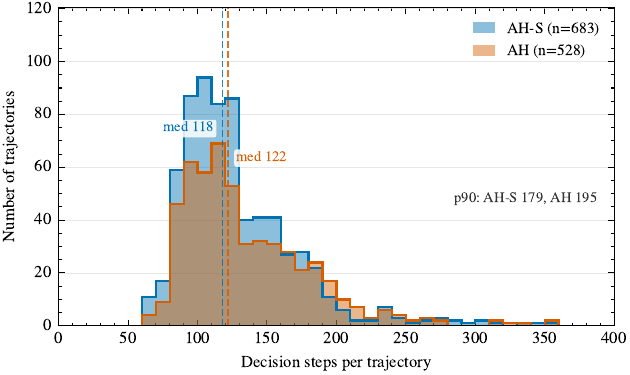}
 \caption{Decision-step distributions for the evaluation subsets. AH has a
 median of $\pnnum{122}$ steps and 90th percentile of $\pnnum{195}$; AH-S has
 $\pnnum{118}$ and $\pnnum{179}$, respectively.}
 \label{fig:steps-hist}
\end{figure}

In Figure~\ref{fig:steps-hist}, we show the distribution of trajectory step counts: the trajectories have a median of $\pnnum{121}$ steps, a 90th percentile of
$\pnnum{186}$, and a maximum of $\pnnum{358}$; $\pnnum{83}$ of the data points have
at least $\pnnum{200}$ steps.

\paragraph{Desktop application statistics}
Our benchmark covers the use of $\pnnum{112}$ desktop applications, with a mean $\pnnum{1.98}$ per
task and a median of 2. $\pnnum{83.2\%}$ of tasks use exactly two applications, indicating that multiple apps are needed to complete long-horizon tasks.

\section{Supplementary results}
\label{app:supp-results}

\subsection{Compute and preprocessing}
\label{app:reproducibility}
\label{app:compute}

\paragraph{Inference}
We use the official APIs of proprietary models, and run open-weight judges via vLLM on H100 GPUs.

\paragraph{Input representation}
For the direct judge approach, we feed one screenshot and action for each of the steps in a trajectory; in order to ensure reasonable context length, we downscale each screenshot to $512{\times}332$ pixels. For open-weight models that cannot support longer contexts, four consecutive frames are merged into a $2{\times}2$ mosaic. Our release
pins the environment versions and provides the exact preprocessing and evaluation scripts to reproduce the experiments.

\begin{table}[t]
\centering
\caption{Resource use for the primary agentic-judge configurations. Means are
computed over all $\pnnum{1{,}373}$ trajectories unless a harness omitted the
corresponding telemetry; self-hosted rows have no marginal API charge.}
\label{tab:cost}
\scriptsize
\begin{tabular}{llrrrr}
\toprule
Model & Interface & Input tokens & Tools & Images & \$/trajectory \\
\midrule
GPT-5.5 & Codex & 513,556 & 18.6 & 11.9 & 0.4035 \\
Gemini 3.1 Pro & Gemini CLI & 1,393,271 & 24.2 & 30.2 & 0.6172 \\
Claude Opus 4.7 & Claude Code & 554,387 & 9.2 & 7.8 & 0.6592 \\
Qwen 3.6 27B & OpenCode & 243,440 & 8.1 & 6.0 & 0 \\
GPT-5.4 mini & Codex & 323,808 & 20.0 & 8.4 & 0.1066 \\
Qwen 3.6 35B-A3B & OpenCode & 190,550 & 8.1 & 5.6 & 0 \\
Claude Haiku 4.5 & Claude Code & 452,266 & 10.6 & 8.3 & 0.1216 \\
Gemini 3.1 Flash Lite & Gemini CLI & 1,041,455 & 8.1 & 3.6 & 0.0949 \\
Gemma 4 26B-A4B & OpenCode & 243,892 & 3.9 & 1.2 & 0 \\
Gemma 4 31B & OpenCode & 101,528 & 2.3 & 1.0 & 0 \\
\bottomrule
\end{tabular}
\end{table}

\subsection{Mistake-type recall}
\label{app:per-mt-frontier}

To supplement results from our binary evaluations, we also include the fine-grained failure-type recall in Table~\ref{tab:per-mt-full}. We found that Bad Side Effect has the lowest recall, indicating models are poorly calibrated to distinguish side effects from other mistake types.

\begin{table}[t]
\centering
\caption{Exact mistake-type recall on the full benchmark for the primary
agentic-judge configurations. Denominators are 273 Critical Mistakes, 232 Bad
Side Effects, and 339 Misunderstandings; malformed outputs count as incorrect.}
\label{tab:per-mt-full}
\scriptsize
\begin{tabular}{llrrr}
\toprule
Model & Interface & Critical & Side effect & Misunderstanding \\
\midrule
GPT-5.5 & Codex & 77.7 & 24.1 & 72.9 \\
Gemini 3.1 Pro & Gemini CLI & 57.5 & 7.3 & 87.3 \\
Claude Opus 4.7 & Claude Code & 75.8 & 12.1 & 71.4 \\
Qwen 3.6 27B & OpenCode & 60.4 & 5.6 & 74.3 \\
GPT-5.4 mini & Codex & 60.8 & 2.6 & 74.6 \\
Qwen 3.6 35B-A3B & OpenCode & 63.0 & 6.0 & 53.1 \\
Claude Haiku 4.5 & Claude Code & 58.6 & 8.6 & 53.4 \\
Gemini 3.1 Flash Lite & Gemini CLI & 62.3 & 1.7 & 29.2 \\
Gemma 4 26B-A4B & OpenCode & 47.6 & 0.0 & 18.6 \\
Gemma 4 31B & OpenCode & 74.0 & 0.0 & 27.1 \\
\bottomrule
\end{tabular}
\end{table}

\subsection{Evaluation with additional harnesses}
\label{app:full-grids}

\label{app:controlled-interface}

In Table~\ref{tab:model-interface-comparison}, we report every model that was evaluated with
more than one agent harness, with GPT-5.5 and Qwen 3.6 27B being evaluated on both Codex and OpenCode, allowing a fair comparison. We found that GPT-5.5's performance is 3.7 percentage points lower when switching from Codex to OpenCode, while Qwen achieves 1.2 percentage points lower balanced accuracy on OpenCode and 20.4 percentage points lower on OpenHands. Based on those results, we can conclude that the choice of harness can substantially affect the performance of the judge, and encourage the use of native interfaces (for example, Codex for the GPT family), or otherwise several open harnesses to understand which one is a better fit.

\begin{table}[t]
\centering
\caption{Models evaluated under multiple agentic interfaces on the final AH and AH-S splits. All runs use prompt P9 and the same benchmark inputs. AH Pos and AH Neg use fixed denominators of 227 and 301; AH-S Bal. uses 234 positives and 449 negatives. Missing or malformed outputs count as incorrect. $^{\dagger}$ marks the primary interface reported in Table~\ref{tab:main-lb-primary}.}
\label{tab:model-interface-comparison}
\small
\begin{tabular}{llrrrr}
\toprule
Model & Interface & AH Bal. & AH Pos & AH Neg & AH-S Bal. \\
\midrule
GPT-5.5 & Codex$^{\dagger}$ & 80.9 & 71.8 & 90.0 & 92.6 \\
 & OpenCode & 77.2 & 70.9 & 83.4 & 92.2 \\
\addlinespace[2pt]
Qwen 3.6 27B & Codex & 71.6 & 70.5 & 72.8 & 94.5 \\
 & OpenCode$^{\dagger}$ & 70.4 & 75.3 & 65.4 & 93.6 \\
 & OpenHands & 51.2 & 58.1 & 44.2 & 89.0 \\
\addlinespace[2pt]
Gemini 3.1 Flash Lite & Gemini CLI$^{\dagger}$ & 52.1 & 58.6 & 45.5 & 75.2 \\
 & OpenCode & 51.3 & 41.9 & 60.8 & 80.5 \\
 & Codex & 47.8 & 38.8 & 56.8 & 73.1 \\
 & OpenHands & 46.1 & 25.1 & 67.1 & 70.9 \\
\bottomrule
\end{tabular}
\end{table}

\subsection{Effect of removing required information}
\label{app:surface-controls}

We study the effect of removing the instruction from the input passed to the judge, which would make this task impossible (since both trajectories should appear valid) in Table~\ref{tab:surface-controls}. We found that removing the instruction and action log results in balanced accuracy close to random (47.8\% to 49.8\%), whereas removing screenshots results in very poor performance (57.7\%). This highlights that the trajectories by themselves do not provide enough information to predict the success, and that screenshots provide important information to make a proper judgment.

\begin{table}[ht]
\centering
\small
\caption{\small{Ablation results evaluating which trajectory components are necessary for judging a trajectory.}}
\label{tab:surface-controls}
\begin{tabular}{ccc rrrr c}
\toprule
\textbf{Instruction} &
\textbf{Screenshots} &
\textbf{Action log} &
\textbf{TP} &
\textbf{FN} &
\textbf{FP} &
\textbf{TN} &
\textbf{BA} \\
\midrule

\xmark &
\cmark\ all &
\cmark &
246 & 5 & 273 & 0 &
\textbf{49.0\%} \\

\cmark &
\textemdash\ final only &
\xmark &
104 & 147 & 118 & 155 &
\textbf{49.1\%} \\

\xmark &
\textemdash\ final only &
\xmark &
215 & 35 & 245 & 26 &
\textbf{47.8\%} \\

\cmark &
\xmark &
\cmark &
205 & 46 & 181 & 92 &
\textbf{57.7\%} \\

\xmark &
\xmark &
\cmark &
249 & 2 & 272 & 1 &
\textbf{49.8\%} \\

\bottomrule
\end{tabular}
\end{table}

\ifsubmission\else
\clearpage
\listoftodos
\fi

\end{document}